# RMA: Context-Orchestrated Research Math Agents

**Zelin Zhao[1], Bo Yuan[1], Yuchen Zhu[1], Jaemoo Choi[1], Yongxin Chen[1]**
[1]Georgia Institute of Technology

Long-horizon mathematical reasoning fails less often because a model cannot produce a valid next step than because an agent fails to maintain and expose the right semantic state across many iterations. Left unmanaged, this produces research-level proofs that are locally convincing yet globally incomplete: a key lemma unproved, an assumption unchecked, a citation unsupported, or a computational claim unverified. We present Research Math Agents (RMA), an agentic framework for long-horizon proof development built around a persistent, typed research store and an orchestrator that compiles operation-specific context from that store. The Research Context Orchestrator is the central state-management layer between the persistent research store and each locally scoped proof operation: it retrieves task-relevant artifacts, compiles them into a bounded context, invokes the appropriate operation, and writes the resulting proof edits, issue updates, literature notes, plans, or evaluations back to the store. This process is designed to keep proof revisions, unresolved issues, prior attempts, literature, and evaluations available across rounds while exposing only task-relevant state to each local operation. We evaluate RMA across complementary research-level settings using independent expert evaluation, blind mathematician review, LLM-based benchmark evaluation, and Lean 4 kernel verification. RMA achieves a 42.5% solve rate on the independently evaluated SOOHAK Challenge Hard set, obtains 8 of 10 correct solutions on First Proof B1 and 8 of 10 passing solutions on B2 under human-expert evaluation, and verifies 213 of 300 sampled Research Solved targets in Formal Conjectures with the Lean 4 kernel. These results support structured state and context orchestration as useful components for long-horizon proof development, while comparisons against independently developed systems remain subject to differences in inference budgets, backbones, and evaluation protocols.

**Main Contact:** zelin@gatech.edu
**Code:** https://github.com/sjtuytc/ResearchMathAgent

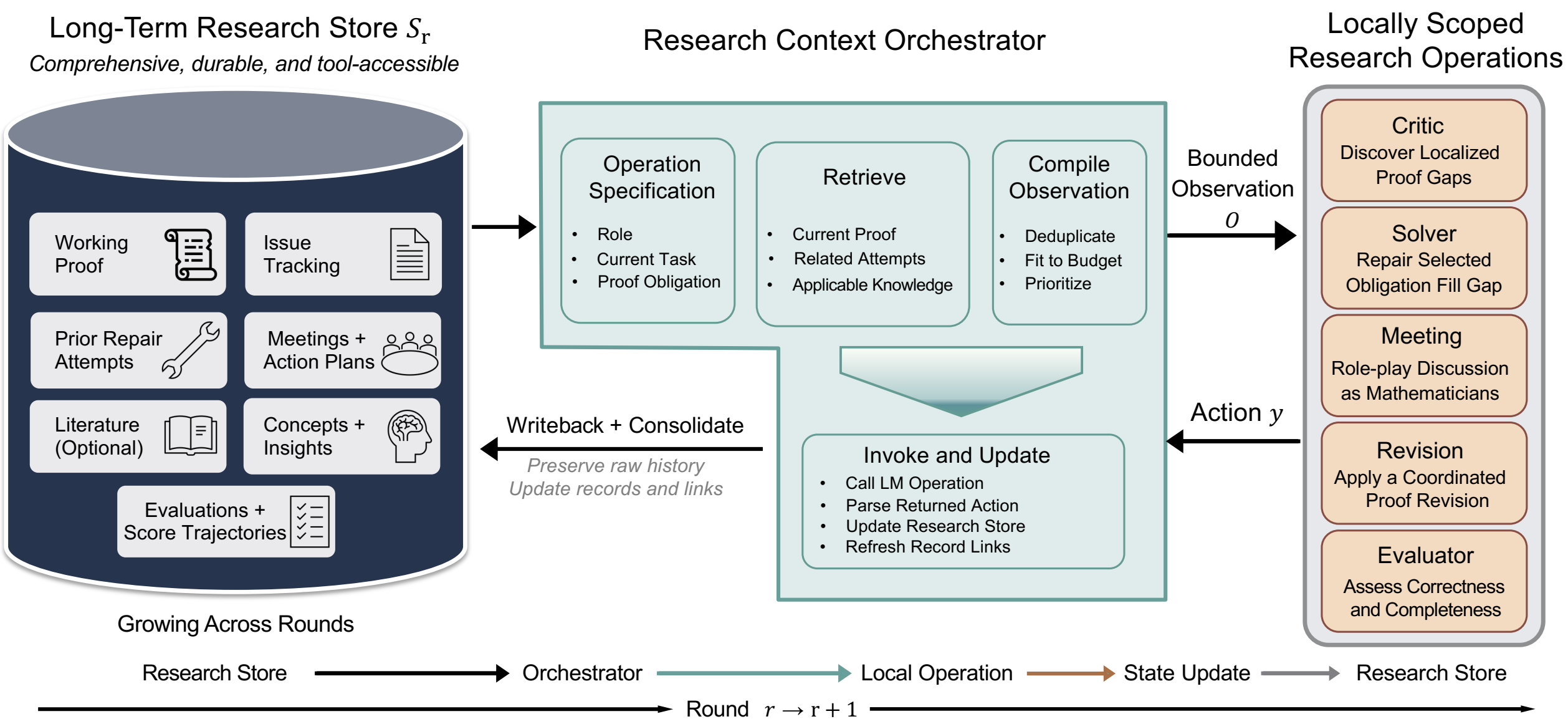

**Figure 1. Overall diagram of RMA.** (**Left**) The research store $\mathcal{S}$ maintains the persistent research state at round $r$, including proofs, issues, prior attempts, meetings, literature, concepts, insights, and evaluations. (**Middle**) The Research Context Orchestrator retrieves and prioritizes relevant artifacts, compiles a bounded operation-specific observation under budget $B$, invokes the corresponding operation, and writes its output back to the store. (**Right**) Locally scoped research operations receive only the compiled observation and return structured artifacts such as proof patches, issue updates, literature notes, plans, insights, or evaluations.

## 1. Introduction

Large language models have made rapid progress on mathematical reasoning, from the MATH dataset [1] to Olympiad problems [2], [3], a domain long viewed as a test of machine intelligence [4] – but concentrated in self-contained settings where a solution fits within a short derivation. Research-level mathematics is different [5], [6]: it requires identifying hidden assumptions, introducing definitions, consulting prior work, maintaining partial progress, and repeatedly deciding which unresolved obligation to address next. Here, strong models often produce proofs that are locally convincing yet globally incomplete – a key lemma unproved, an assumption unchecked, a citation unsupported, a computational claim unverified [7]

– failing by *omission*: reading as finished even though a required part of the argument remains open.

Addressing omission requires more than generating additional candidates or asking a model to critique the current proof. Repairing a missing step depends on a structured research context that records the unresolved claim, its dependencies, prior attempts, relevant literature, and the parts of the argument already established. Existing LLM agents provide planning, memory, critique, and proposer–verifier interaction [8], [9], [10], but they often operate over generic conversation histories and invoke broad actions such as proposing, debating, or revising. Such operations may detect a gap without retaining the evidence, dependencies, and failed attempts needed to resolve it, so later revisions can improve the presentation while leaving the same proof obligation open. Formal theorem provers avoid this ambiguity because proof assistants expose explicit subgoals [11], [12]; step-level methods for informal reasoning can score individual steps [13] or obtain localization from an external formal solver [14], but they do not maintain the evolving research state required for long-horizon natural-language proof development. We instead organize proof development around a persistent research store and locally scoped operations, so that each operation receives the specific context required to identify, prioritize, or repair an unresolved proof obligation.

We take the position that a major source of failure in long-horizon mathematical reasoning is *state management*: even when a model can produce useful local steps, an agent may repeatedly re-derive, forget, or silently drop unresolved obligations if it does not preserve what remains open, what those obligations depend on, and what has already been tried. This motivates RMA, a stateful architecture in which global research state persists across local operations, while each operation receives only the task-relevant context compiled from that state and writes its result back afterward (Figure 1). Concretely, RMA coordinates established reasoning operations through three mechanisms: (i) a typed Research Store spanning proof revisions, issues, attempts, literature, concepts, meetings, insights, and evaluations; (ii) severity-and-dependency-based prioritization of unresolved proof obligations; and (iii) operation-specific context construction through the Research Context Orchestrator. Section 3 details these mechanisms, and Section 4.2 isolates their contribution experimentally.

The central question we study is whether explicitly structuring persistent proof state and selectively compiling it into local observations improves long-horizon mathematical reasoning under a common inference-budget constraint. In **R**esearch **M**ath **A**gents (RMA), each push-forward round discovers issues through an LM critic, structural claim-dependency checks, a semantic completeness judgment, and a fidelity analysis; prioritizes them by severity and dependency impact; repairs a bounded subset; incorporates relevant literature and coordinated revision; and evaluates the updated proof. When the full claim cannot be established, reduction mode records the strongest complete partial result while leaving the remaining obligation explicitly open.

We evaluate RMA across four complementary research-level settings. **First**, on the independently evaluated SOOHAK Challenge Hard set [15], RMA achieves a 42.5% solve rate, compared with 35.0% for GPT-5.5 and 12.5% for its standalone Claude Opus 4.8 backbone. **Second**, under human-expert evaluation on First Proof [7], [16], RMA obtains 8 of 10 correct solutions on B1 (with a Claude Opus 4.6 backbone; Section 4.1) and 8 of 10 passing solutions on B2. **Third**, on ResearchMath [17], it obtains a 5.6% positive lemma-decomposition rate on the reported 90-problem evaluation sample. **Fourth**, on Formal Conjectures [18], RMA verifies 213 of 300 sampled Research Solved targets with the Lean 4 kernel, compared with 193 for Seed-Prover [19] and 121 for direct Claude Opus 4.8 on the same set.

RMA makes three **contributions**:

- *Persistent Research Store*: a typed state schema that preserves proof revisions, issue records, prior attempts, literature, meetings, concepts, insights, and evaluations across research rounds, rather than treating them as an undifferentiated conversation history or flat fact log.
- *Research Context Orchestrator*: severity-and-dependency issue prioritization together with operation-specific retrieval and context construction, so critic, solver, literature, meeting, revision, and evaluation calls receive different task-relevant views of the persistent research state.
- *Broad empirical evaluation*: evaluation across SOOHAK Challenge Hard [15], First Proof B1/B2 [7], [16], ResearchMath [17], and Formal Conjectures [18], spanning independent benchmark evaluation, human mathematical review, LLM-based benchmark judging, and Lean 4 kernel verification.

## 2. Related Work

**Research-level mathematical reasoning and theorem proving.**
Mathematical-reasoning benchmarks have progressed from arithmetic and competition settings—including GSM8K, MATH, OlympiadBench, and OmniMath [1], [20], [21], [22]—toward research-level and open-ended mathematical tasks. First Proof [7] is particularly close to our setting, requiring the development and evaluation of research-level proofs rather than short contest-style derivations. Formal and neural theorem provers provide machine-checked guarantees [23], [24], [25], [26]. Earlier stateful systems such as COPRA [27] and ReProver [28] organize search around formal tactic states, while more recent systems including NanoProof [29], Seed-Prover [19], and Hilbert [30] substantially strengthen formal proof generation through learned and test-time reasoning pipelines. We include representative recent systems in our same-target Formal Conjectures evaluation. RMA differs in scope: its persistent research state and context orchestration are designed to support both informal research-level proof development and formally verified proof generation.

**Agentic reasoning and autonomous research.** A broad line of work augments language models with verifier-guided generation, critique and revision, tool use, planning, and multi-agent interaction [10], [13], [31]. Deep-research assistants [32], [33] and autonomous-research systems [6], [34] extend these ideas to longer-horizon information gathering and iterative synthesis. RMA shares this iterative-agent perspective but specializes the execution state to mathematical proof development: critic, solver, literature, meeting, revision, and evaluation operations communicate through persistent proof-specific records rather than only through a conversational history or successive drafts.

**Context engineering and persistent memory.** RMA is also closely related to work on external memory and context engineering. MemGPT [35] moves information between a bounded active context and an external store, while N. F. Liu *et al.* [36] show that language models use information non-uniformly across long contexts, motivating selective context construction. ReasoningBank [37] stores and retrieves reusable reasoning strategies, and blackboard-style architectures coordinate multiple agents through a shared workspace [38]. RMA adopts the broader principle of persistent external state but specializes it to long-horizon proof development: proof revisions, issue records carrying severity and dependency metadata, prior attempts, literature, meetings, concepts, strategic insights, and evaluations are represented separately and selectively exposed to later operations.

## 3. Methodology

As summarized in Figure 1, RMA has three layers: the research store maintains the persistent state of proof development; the Research Context Orchestrator converts this global state into bounded interactions and coordinates state transitions; and locally scoped research operations instantiate the system as low-level LM calls, each receiving a bounded, operation-specific observation and returning a structured artifact for integration into the research store. We discuss these three parts in the following subsections, in order, and give the full end-to-end pseudocode as Algorithm 1 in Appendix D, where it is annotated line by line.

### 3.1. Persistent Research Store

We organize proof development into *research rounds*. A round is one outer iteration in which the system examines the current proof, identifies and prioritizes unresolved obligations, performs a bounded collection of research and repair operations, revises the proof, and evaluates the resulting state. We denote the total round budget by $N_R$ and index rounds by $r \in \{0, \ldots, N_R - 1\}$. Execution may stop before exhausting this budget when the proof is judged complete or the system ceases to make measurable progress across two consecutive rounds.

The *research store* $\mathcal{S}$ is the persistent research state maintained across rounds, and each question has one research store $\mathcal{S}$. We define it as $\mathcal{S} = (\Pi, \mathcal{I}, \mathcal{M}, \mathcal{L}, \mathcal{K}, \mathcal{H}, \mathcal{E})$, where $\Pi$ contains proof artifacts and their revision history, including the current proof $\pi$; $\mathcal{I}$ contains open and resolved proof issues and obligations; $\mathcal{M}$ contains records of structured research meetings in which role-specialized agents simulate collaboration among mathematicians and synthesize actionable plans; $\mathcal{L}$ contains literature records and associated notes when literature grounding is enabled, and may remain empty when the literature module is disabled; $\mathcal{K}$ contains mathematical concepts and background knowledge; $\mathcal{H}$ contains strategic insights, which describe potential uncovered directions for solving the problem; and $\mathcal{E}$ contains evaluation records. At the beginning of the algorithm, we initialize the research store $\mathcal{S}$ from the problem $p$, including the initial proof state, empty issue and evaluation records, and the persistent structures used to accumulate literature, concepts, meeting outcomes, and insights.

### 3.2. Research Context Orchestration

**Motivation and definition.** Although the research store preserves the state required for long-horizon proof development, an individual LM call should not receive the entire store: context windows are bounded, and models do not reliably use all information in long inputs [36]. The *Research Context Orchestrator* therefore serves as the central execution and state-management layer between $\mathcal{S}$ and the locally scoped operations, determining what a task needs, retrieving and compiling it into a focused, budget-satisfying observation, invoking the operation, and integrating the result back into the research state. It takes as input a problem $p$, a total round budget $N_R$, a per-round issue budget $b$, and a per-call context budget $B$, all fixed before the run begins.

**Context compilation.** $\mathrm{PrefixToBudget}(C, B)$ operates on $C$ as a sequence of whole, addressable *records*, never on raw characters or tokens, and removes them strictly in ascending priority order until the serialized observation $O$ fits within $B$, so an operation never receives a lemma statement or proof fragment cut mid-sentence. $\mathrm{LinkedRecords}(\mathcal{S}, q)$ and $\mathrm{RecentOutputs}(\mathcal{S}, u)$ are themselves query- and operation-specific retrieval functions, not blanket reads of $\mathcal{S}$; $\mathrm{CurrentProof}(\mathcal{S})$ returns the latest full proof revision, not a sub-document. The priority order is the concatenation order of line 20 in Algorithm 1, so records are dropped from the end first.

**Issue discovery.** Long-horizon proofs may contain locally plausible arguments while leaving crux lemmas, cita-tion-dependent steps, computational claims, or other proof obligations unresolved [7]. The critic operation applies four complementary analyses to the current proof. First, an LM-based critic identifies logical gaps and returns them as typed issue records. Second, a deterministic structural checker parses the proof into claims and dependencies, measures the fraction of terminal claims supported by completed arguments, and flags finite or numerical claims that lack executable verification. Third, a semantic completeness judge assesses whether the proof establishes the target statement as a whole and, when its score falls below the completeness threshold, converts its itemized missing steps into severity-rated issues. Those three all grade the proof *given* whatever reading of the problem it adopted, so a fourth, independent *fidelity* analysis asks whether that reading is the intended one or one that trivializes the problem relative to its own citations — a failure an internally airtight proof can pass. Each issue records the affected claim, defect type, missing justification or evidence, required repair, and whether the gap is correctness-critical —that is, whether the target result depends on closing it. This criticality flag is used by severity assignment and termination, so issue type alone does not determine whether a proof may be declared complete.

**Prioritized issue repair.** Each issue receives a severity according to its effect on proof validity rather than its defect type alone. P0 denotes a defect that invalidates the main conclusion or leaves a necessary crux claim unsupported; P1 denotes a gap that blocks a required lemma or proof branch; P2 denotes a localized missing justification, citation detail, or verification whose absence does not invalidate a necessary step; and P3 denotes a notation or presentation defect that does not affect correctness. In particular, a citation-dependent step is not automatically P2: a `cited_blackbox_crux` marked critical by the gap detector is promoted to correctness-critical severity and must be closed before the proof can be declared solved. The critic orders open issues first by severity, then by dependency impact – the number of unresolved downstream claims depending on the affected claim – both computed deterministically from the critic's criticality flags and a claim-dependency graph extracted from the proof text and rebuilt every round (Appendix D gives the extraction mechanism). The following loop restricts repair to the first $b$ issues: for each selected issue $\iota$, the solver receives an observation with the current proof, issue specification, dependencies, related attempts, and applicable knowledge, and returns a localized edit, auxiliary lemma, citation replacement, or explicit verification – an anchored patch (exact span plus replacement) applied only if the anchor is found unambiguously and disturbs no outside text, with a full-document-rewrite fallback (logged, gated by a regression check, Appendix E) after one failed retry. When the complete claim cannot be established, reduction mode instead records the strongest complete sub-result available and explicitly leaves the remaining proof obligation open for a later round.

**Issue resolution and revalidation.** An issue is marked resolved when the solver's patch or rewrite is written back. Rather than immediately rechecking that specific issue, the next critic round re-scans the full current proof; any remaining or reintroduced defect is then recorded as a new issue. Appendix D gives the corresponding deduplication rule.

**Optional literature grounding.** Literature grounding is an optional module rather than a requirement of the RMA architecture. When enabled, the orchestrator invokes the Literature operation with the problem $p$ and current ranked issue set $\mathcal{Q}$ to identify references relevant to unresolved claims by mathematical objects, theorem dependencies, assumptions, and equivalent formulations. For each useful source, it records the relevant theorem or technique, its assumptions, the claims it may support, and conditions limiting applicability, and persists these records in $\mathcal{L}$. When external retrieval is disabled, web search and fetch are unavailable to the agent and the remainder of the RMA pipeline operates without externally retrieved literature. Thus, persistent state, issue prioritization, local repair, coordinated revision, and evaluation do not require network access. When literature grounding is enabled, Literature runs after the current round's solver loop has consumed $\mathcal{Q}$ (Algorithm 1); consequently, sources retrieved from round-$r$ issues become available to the solver of the next round.

**Meetings, coordinated revision, and concept tracking.** Local issue repair can become myopic, so the meeting operation convenes a coordinator and field-appropriate mathematician personas over the current proof, issues, attempts, and literature, synthesizing a meeting record $\rho$, action plan $a$, and strategic insights $\Delta\mathcal{H}$ that the revision operation applies as a coordinated proof update, advancing proof development through both localized repairs and broader strategic revisions; Appendix D summarizes the persona and synthesis protocol. Alongside this, the concept component $\mathcal{K}$ is a structured, problem-specific glossary of the proof's mathematical objects – name, definition, assumptions, notation, source – that standardizes terminology across the research history and any available literature records, initialized from the problem statement and updated after each revision using the current proof and meeting insights; Appendix D gives the entry schema and update rule in full.

**Evaluation, round finalization, and termination.** The evaluator operation assesses the current proof on answer accuracy, logical correctness, completeness, and clarity; its output contains both scores and a written assessment and is appended to the evaluation component $\mathcal{E}$. We additionally record the proved-leaf fraction, completeness score, and the issues opened and closed in each round, making progress in proof completion directly measurable (fur-ther details in Appendix D). FinalizeRound then consolidates the round's writes into $\mathcal{S}$; the loop terminates through $\mathrm{Solved}(\pi, e)$ only when the semantic completeness gate passes and no open issue is marked correctness-critical. Correctness-critical issues include any gap whose absence would invalidate the target conclusion or a necessary supporting claim, regardless of issue type; thus an unresolved critical `cited_blackbox_crux`, unproved lemma, logical leap, or unchecked computational claim blocks Solved. The loop may otherwise terminate only through $\mathrm{Stalled}(\mathcal{S})$ or exhaustion of $N_R$.

### 3.3. Locally Scoped Research Operations

As shown in the right of Figure 1, the orchestration layer is implemented through a collection of locally scoped research operations, such as the critic, solver, optional literature researcher, meeting, reviser, and evaluator as we introduced before. Each operation is responsible for one explicit research task and receives only the bounded, operation-specific observation prepared by the orchestrator, rather than direct access to the complete research store $\mathcal{S}$. It returns a structured artifact—such as an issue update, proof patch, literature note, action plan, strategic insight, revised proof, or evaluation—that the orchestrator validates and writes back to the corresponding component of $\mathcal{S}$. This common interface keeps the operations modular while allowing their outputs to accumulate into a single persistent, auditable record of proof development over time.

**Run function.** Every operation executes through the shared procedure $\mathrm{Run}(u, q, \mathcal{S}, B)$, where $u$ is the operation and $q$ its local task—the problem alone, or paired with a selected issue, issue queue, action plan, or current proof. Run compiles the fixed-priority context $C$, truncates it to a bounded observation $O$ via PrefixToBudget (detailed above), invokes $u$ on $O$ alone rather than on the complete store, and writes the resulting artifact $y$ back into the corresponding components of $\mathcal{S}$; no operation is purely LM-based or purely deterministic, since even the LM-generated artifacts pass through a deterministic write-back gate. Appendix D gives the full pseudocode (Algorithm 1) and, in Table 5, an operation-by-operation breakdown of the mechanisms that produce each artifact.

## 4. Experiments

**Benchmark suites and evaluation schemes.** (**a**) **SOOHAK Challenge Hard** [15]: 40 non-public, contamination-resistant graduate/research-adjacent problems; we submitted RMA's API endpoint to the organizers, who independently ran three seeded repetitions for RMA and the reported baselines under their official expert protocol. Seeds, execution, and grading were controlled by the organizers, making this a **third-party evaluation**; the comparison is protocol-matched but not compute-matched across systems. (**b**) **First Proof** B1 [7]/B2 [16]: ten research problems each from expert mathematicians; ten mathematicians (at least three per problem) conduct blinded correctness evaluation (protocol in Appendix A; leakage controls in Appendix B). (**c**) **ResearchMath** [17]: the authors' own 90-problem sample and LLM-judge protocol; a disjoint held-out set of 30 problems the benchmark annotates as solved is used only for the scalability and ablation studies (Appendix C). (**d**) **Formal Conjectures** [18]: Lean 4 [23]-formalized conjectures. For the quantitative evaluation reported in this paper, we restrict to the benchmark's *research* category (excluding textbook/API-stub/build entries) and uniformly sample 300 *Research Solved* targets with a fixed dataset-sampling seed and no AI-solvability filtering. All methods in Table 2 are evaluated on these same 300 target IDs under the corresponding pinned Lean 4/Mathlib environments. Exploratory Research Open runs are excluded from our quantitative claims; we explain this scope choice in Appendix G. Sampling for the other three benchmarks is fixed by their own organizers, not by us. The default backbone is Claude Opus 4.8, except for First Proof B1, which was run earlier with Claude Opus 4.6 [39] (Appendix G); total recorded API spend across the experimental runs was approximately $50,000 USD.

### 4.1. Comparison Results with Prior Works

**SOOHAK Challenge Hard [15].** Table 1 reports an independent third-party evaluation conducted entirely by the SOOHAK organizers under their official expert protocol. The organizers evaluated every reported system, including RMA and all baselines, over three seeded repetitions of the same 40-problem benchmark; the reported solve rates are averages over those organizer-run repetitions. Thus, the repeated runs and their seeds were neither selected nor executed by us. All RMA runs use the default `claude-opus-4-8` backbone.

> **Takeaway 1.** Under the SOOHAK organizers' three-repetition evaluation, RMA achieves a mean solve rate of **42.5%**, compared with **35.0%** for GPT-5.5 and **12.5%** for its Claude Opus 4.8 backbone. All systems are evaluated by the organizers under the same benchmark protocol.

**First Proof B1 and B2 [7], [16].** Table 1 (right) and Table 2 (left) evaluate RMA on the two First Proof batches under their respective human-expert protocols. For B1, both RMA and its standalone Claude baseline use the earlier Claude Opus 4.6 backbone; B2 and the later experiments use Opus 4.8 (Appendix G). We compare B1 against the publicly released First Proof submissions from GPT-5.2R [40] and Aletheia [41], together with OpenAI Deep Research (GPT-DR) [32] and Gemini Deep Research (Gemini-DR) [33], all evaluated under the same blinded expert protocol. For B2 we follow the comparison set adopted by its authors – ProofCouncil [42], UCLA Moonshot [43], ChatGPT 5.5 Pro [44], and Princeton Momus [45] – and also report total run cost. The headline RMA run costs $1,850,

**Table 1. Human-expert evaluations (NOT LLM-judged).** Both panels are graded by human mathematicians. *Left:* third-party evaluation on SOOHAK Challenge Hard [15]. The benchmark organizers independently ran RMA and every reported baseline for three seeded repetitions under the same official 40-problem expert-evaluation protocol; solve rates report the mean across those repetitions. Seeds, execution, and grading were controlled by the organizers, not by us. *Right:* blind expert verdicts on First Proof B1 [7]: correct (✓), inconclusive (≈), and incorrect (×); ✓/× require unanimity among at least three experts, and ≈ is recorded whenever they disagree. †Aletheia released solutions for only 6 of the 10 problems: 5 were unanimously judged correct and 1 inconclusive; the remaining 4 problems had no released solution (N/A).

| **SOOHAK Challenge Hard** — 3rd-party expert | | |
|---|---|---|
| Method | Solve (%) ↑ | Runs |
| **RMA** (Opus 4.8) | **42.5** | 3 |
| GPT-5.5 | 35.0 | 3 |
| Gemini 3.1 Pro | 17.5 | 3 |
| Claude Opus 4.8 | 12.5 | 3 |
| Kimi K2.7 Code | 7.5 | 3 |
| DeepSeek-V4-Flash | 5.0 | 3 |

| **First Proof B1** — blind human expert | | | |
|---|---|---|---|
| Method | ✓ ↑ | ≈ | × ↓ |
| **RMA** (Opus 4.6) | **8** | 1 | 1 |
| Aletheia† | 5 | 1 | N/A |
| GPT-5.2R | 3 | 4 | 3 |
| Opus 4.6 | 0 | 1 | 9 |
| GPT-DR | 0 | 0 | 10 |
| Gemini-DR | 0 | 0 | 10 |

**Table 2. Human-expert Pass/Fail and Lean 4 kernel verification (NOT LLM-judged).** *Left:* First Proof B2 [16] submissions graded by human mathematicians as **Pass** (flawless or needing only minor revisions) or **Fail**, with total run cost. *Right:* Lean 4-kernel verification on the same 300 sampled Formal Conjectures Research Solved targets. We evaluate RMA, its direct Claude Opus 4.8 backbone, NanoProof [29], Seed-Prover [19], and Hilbert [30] on the same target IDs and pinned Lean 4/Mathlib environments.

| **First Proof B2** — blind human expert | | | |
|---|---|---|---|
| Method | Pass ↑ | Fail | Total cost ($) |
| **RMA** (Opus 4.8) | **8** | 2 | 1,850 |
| ProofCouncil | 7 | 3 | 3,186 |
| UCLA Moonshot | 3 | 7 | 4,799 |
| ChatGPT 5.5 Pro | 3 | 7 | **117** |
| Princeton Momus | 1 | 9 | 1,014 |

| **Formal Conjectures Research Solved** — Lean 4 | | |
|---|---|---|
| Method | Setting | Verified ↑ |
| **RMA** (Opus 4.8) | agentic | **213/300** |
| NanoProof | prover | 135/300 |
| Seed-Prover | prover | 193/300 |
| Hilbert | prover | 164/300 |
| Opus 4.8 direct | direct | 121/300 |

below ProofCouncil ($3,186) and UCLA Moonshot ($4,799), though substantially above ChatGPT 5.5 Pro ($117).

> **Takeaway 2.** RMA obtains 8 of 10 correct B1 solutions and 8 of 10 passing B2 submissions. The headline B2 run realizes $1,850 in total API spend; a separate higher-budget run produces no additional pass.

**Formal Conjectures [18] (Lean 4 kernel, not LLM-judged).**
Table 2 (right) reports Lean 4-kernel verification on the same 300 sampled Research Solved targets. With external web search and fetch disabled, RMA verifies 213/300 (71.0%), compared with 193/300 (64.3%) for Seed-Prover, 164/300 (54.7%) for Hilbert, 135/300 (45.0%) for NanoProof [29], and 121/300 (40.3%) for direct Claude Opus 4.8. The direct-backbone result provides a same-backbone reference without the Research Store or orchestration layer, while the specialized prover results provide stronger contemporary cross-system comparisons. Because the specialized systems differ in backbone and test-time inference procedure, we treat these comparisons as descriptive rather than matched-compute controls. Every reported RMA success is accepted by the Lean 4 kernel under the benchmark's pinned environment; we reject proofs containing `sorry` or `admit`, depending on `sorryAx` or project-local custom axioms, or using `native_decide` to bypass the target proof.

> **Takeaway 3.** On Formal Conjectures Research Solved, RMA verifies **213/300** targets with the Lean 4 kernel, compared with **193/300** for Seed-Prover and **121/300** for direct Claude Opus 4.8.

**ResearchMath [17] (complementary LLM-judge evaluation).**
Table 3 (left) scores whether traces explicitly decompose research-level problems into useful intermediate lemmas under the benchmark's own LLM-judge protocol. ResearchMath provides a complementary LLM-judged measure of lemma decomposition: RMA obtains 5/90 positives versus 3/90 for its direct Opus 4.8 backbone and 3/90 for Qwen3.5-397B. This quantity is not directly comparable to the 25/30 human pass count on the held-out ablation set: the former measures positive lemma decomposition on the benchmark's 90-problem sample, whereas the latter evaluates complete proofs on a separate subset restricted to questions annotated as solved.

## 4.2. Cost, Scalability, and Ablation Studies

**Cost and scalability.** Figure 2(a) reports separate budget-scaling runs over the same ten First Proof B2 problems. The $1,850 run reaches 8/10 passes and is the headline RMA result reported in Table 2. A separate higher-budget run realizes $3,571 while reaching the same 8/10 pass count, so increasing spend beyond $1,850 produces no additional pass over the tested budget points. Thus, over the tested budget points, the observed B2 pass count reaches its maximum by the $1,850 headline run and remains unchanged at the higher $3,571 spend. The studies below instead use 30 held-out solved ResearchMath problems, disjoint from every problem used in our earlier runs, under a common $100-per-problem monetary cap (Appendix C). Reading off what a run capped at $N_R = r$ would deliver (Figure 2b; construction in Appendix C), the judge score

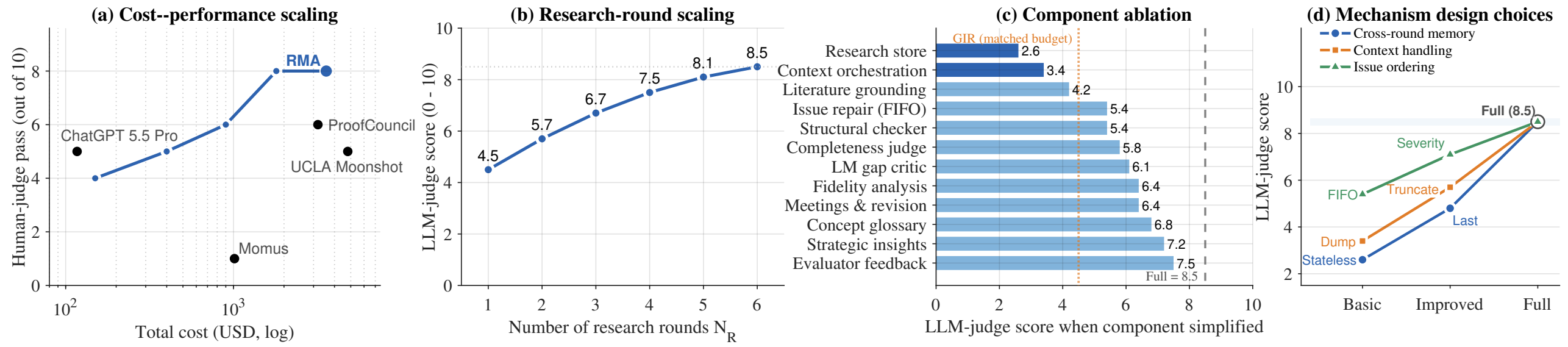

**Figure 2. Cost, scalability, and ablations of RMA.** (**a**) Human-expert passes versus realized total API spend on First Proof B2 [16]; each RMA point is a separate run under a fixed budget, plotted at what it actually spent, and all systems are graded by our invited panel. (**b**)–(**d**) Continuous LLM-judge score on the held-out 30-problem ResearchMath ablation set (Appendix C); every configuration, including Generic Iterative Revision, runs under the same hard budget of \$100 per problem: (**b**) as the number of research rounds $N_R$ increases; (**c**) with each component reduced to its simplest functioning variant (dashed line: full system; dotted: Generic Iterative Revision at the same budget); (**d**) for graded design choices in memory, context handling, and issue ordering.

rises from 4.5 to 8.5 over six rounds with per-round gains from 1.2 to 0.4.

**Matched-harness memory ablation.** We evaluate the contribution of structured persistent memory on the same 30-problem held-out ResearchMath set using a matched-harness ablation. The RMA reasoning harness is held fixed, while the representation and retrieval of persistent cross-round memory are changed: full RMA maintains typed, individually addressable persistent records with record-level retrieval, whereas the summary-memory variant represents persistent research state in summarized form. The backbone, prompts, operations, tools, issue-ranking procedure, stopping rule, selection policy, context budget, and \$100 per-problem inference budget are otherwise unchanged. Over three repeated runs, full RMA scores $8.53 \pm 0.15$ under the continuous judge, compared with $6.63 \pm 0.15$ for the summary-memory variant, a 1.90 -point mean gap. Separately, blinded mathematician evaluation of the 30 paired outputs yields 25/30 passes for full RMA versus 16/30 for summary-memory RMA. Among the 17 discordant problem pairs, full RMA passes in 13 cases and the summary-memory variant in 4 (two-sided exact McNemar $p = 0.049$; Table 3). The continuous scores average three repeated runs. The human evaluation is a separate paired evaluation of 30 full-RMA and 30 summary-memory outputs; it is not an aggregation across repeated runs. Accordingly, the 25/30 versus 16/30 pass counts and the McNemar test are computed over 30 paired problems, whereas $8.53 \pm 0.15$ and $6.63 \pm 0.15$ summarize repeated continuous-judge runs. We use this experiment specifically as a matched-harness ablation of persistent-memory representation and retrieval, rather than as a comparison between independently designed systems.

**Broader architectural control.** Generic Iterative Revision (GIR) replaces the structured research-state and operation pipeline with a generic critique-and-revise loop. On the same held-out ResearchMath set, GIR scores 4.5 under its default output-selection rule and $5.50 \pm 0.10$ when given RMA's selection rule. Separately, on First Proof B2 under the same \$1,850 total API budget as the headline RMA run, GIR passes 4/10 problems versus 8/10 for RMA under the blind expert protocol.

**Component ablations.** Figure 2(c) measures the sensitivity of RMA to weakening individual mechanisms on the 30-problem held-out ResearchMath set. The largest degradations occur when the system loses the mechanisms that preserve and expose research state: replacing the persistent store with a stateless variant reduces the score from 8.5 to 2.6, while replacing bounded context construction with direct context dumping reduces it to 3.4. Disabling structured literature grounding gives 4.2, unordered FIFO issue repair gives 5.4, and removing meetings and coordinated revision gives 6.4. The critic analyses contribute more incrementally: disabling the structural checker, semantic completeness judge, LM gap critic, and fidelity analysis yields 5.4, 5.8, 6.1, and 6.4, respectively. Removing the concept glossary, strategic insights, or evaluator feedback has smaller but still measurable effects, yielding 6.8, 7.2, and 7.5. These single-component results therefore place persistent state and bounded context construction among the most consequential mechanisms in the full system, while the remaining components provide additional gains rather than a single dominant source of performance (Appendix C.2).

**Graded design choices.** Panel (d) tests whether the same pattern persists under milder interventions rather than only under the weakest runnable variants. Cross-round memory improves monotonically from stateless (2.6), to retaining only the previous round (4.8), to full persistent state (8.5). Context handling similarly improves from dumping the complete state (3.4), to naive truncation to the context budget (5.7), to operation-specific context compilation (8.5). Finally, issue ordering improves from FIFO (5.4), to severity-only prioritization (7.1), to severity-plus-dependency ranking (8.5). These graded comparisons show that the gains are not limited to catastrophic component removal: retaining some memory is better than none, fitting context to the window is better than dumping it, and severity information improves over unordered repair, while the full mechanisms provide additional gains in each case. The repeated decisive configurations are consistent with the plotted ordering: full RMA scores $8.53 \pm 0.15$, severity-only ordering $7.23 \pm 0.15$, context dumping $3.47 \pm 0.21$, and the stateless variant $2.63 \pm 0.25$ (Appendix C.3). Two interventions require qualification. Context dumping changes request size and completion behavior as well as information selection, so its drop is a joint information-and-resource effect rather than a pure measure of irrelevant context. Likewise, disabling the literature operation removes structured literature retrieval and persistence rather than all web access, because the remaining operations retain their web tools (Appendix C.2).

**Table 3. ResearchMath and matched-harness memory ablation.** *Left:* positive lemma-decomposition count and rate on the ResearchMath [17] 90-problem sample under its benchmark LLM-judge protocol. *Right:* matched-harness memory ablation on a disjoint 30-problem held-out ResearchMath set. The backbone, prompts, operations, tools, issue ranking, stopping rule, selection policy, and per-problem budget are held fixed; only the typed Research Store and record-level retrieval are replaced by summary memory. The LLM score reports the mean and standard deviation over three repeated runs; the human column separately reports blinded mathematician Pass/Fail judgments on one paired set of 30 full-RMA and summary-memory outputs. Among the 17 discordant problem pairs, full RMA passes in 13 cases and summary memory in 4 (two-sided exact McNemar $p = 0.049$).

| **ResearchMath** — LLM judge | | |
|---|---|---|
| Model | Pos. ↑ | Rate (%) ↑ |
| **RMA** (Opus 4.8) | **5/90** | **5.6** |
| Opus 4.8 | 3/90 | 3.3 |
| Qwen3.5-397B | 3/90 | 3.3 |
| DeepSeek V4-Pro | 2/90 | 2.2 |

| **Matched-harness memory ablation** — 30 held-out problems | | |
|---|---|---|
| Configuration | LLM score ↑ | Human Pass ↑ |
| **Full RMA** | $\mathbf{8.53 \pm 0.15}$ | **25/30** |
| Summary memory | $6.63 \pm 0.15$ | 16/30 |

**Interpretation and scope.** We use these ablations as mechanism-sensitivity analyses under a fixed $100-per-problem regime, rather than as estimates of deployment-scale effect sizes or as statistical rankings among nearby variants. In particular, we do not base conclusions on small differences between single-run configurations; the repeated configurations are used to check the larger effects highlighted above. The central claim about structured persistent state also does not rely on the LLM judge alone: the matched-harness comparison on the same held-out set is independently evaluated by blinded mathematicians, yielding 25/30 passes for full RMA versus 16/30 for summary memory, with 13 versus 4 discordant outcomes. We therefore interpret panels (c)–(d) as mechanistic evidence that complements, rather than substitutes for, the human-evaluated matched-harness control.

## 5. Conclusion

We introduced RMA, a context-orchestrated framework for research-level mathematics that maintains structured persistent state across bounded, locally scoped proof operations. On a disjoint held-out ResearchMath set, replacing the typed Research Store and record-level retrieval with summary memory reduces blinded mathematician passes from 25/30 to 16/30 (two-sided exact McNemar $p = 0.049$), supporting the value of structured persistent state. Beyond this controlled ablation, RMA achieves 42.5% on SOOHAK Challenge Hard, obtains 8/10 correct solutions on First Proof B1 and 8/10 passing solutions on B2, and verifies 213/300 sampled Research Solved targets in Formal Conjectures with the Lean 4 kernel.

## AI Use Statement

We disclose the following uses of AI tools in the preparation of this paper. *To aid or polish writing:* standard AI writing tools were used to improve grammar and style in portions of the text; all technical content was authored and verified by the authors. *For retrieval and discovery:* AI-assisted search was used to help locate candidate related work during literature review, with every citation independently checked against its original source before inclusion. *For proving mathematical claims:* RMA, the research-math-agent system that this paper studies, generated the candidate proofs reported and evaluated throughout the paper; every reported result was evaluated under the human-expert, LLM-judge, Lean 4-kernel, or combined verification protocol described for the corresponding benchmark (Section 4), and no AI-generated proof is presented anywhere as a verified result without passing the corresponding protocol.

## Ethics Statement

RMA is designed as a tool for assisting mathematical research, not as a replacement for expert judgment. Used appropriately, it can help researchers explore proof strategies, organize relevant literature, test intermediate arguments, and turn partial ideas into more coherent proof drafts. These benefits are especially relevant for research-level mathematics, where progress often depends on long-horizon reasoning, literature grounding, and repeated refinement rather than one-shot problem solving [5], [6], [7], [46]. The modular design of RMA also makes the reasoning process more inspectable than a single generated answer: problem analysis, retrieved results, proof states, and verifier feedback are stored separately, making it easier for users to audit where a claim comes from.

The main risk is that an incorrect proof may look convincing. This is a known failure mode of LLM-based reasoning systems, where fluent explanations can hide missing assumptions, invalid deductions, or hallucinated references [10], [47]. In mathematics, this risk is particularly serious because a single unjustified lemma can invalidate an otherwise plausible argument. For this reason, RMA should be used in a human-in-the-loop setting: expert mathematicians remain responsible for checking assumptions, verifying citations, and deciding whether a proof is correct. Our use of verifier agents, assumption checks, and expert evaluation is intended to reduce these risks, but it does not eliminate the need for human review. The blind natural-language expert judgments reported here are distinct from formal verification: they reflect the considered opinion of domain experts under time constraints, not a machine-checked guarantee. Any downstream use of RMA's natural-language output should treat it as a candidate requiring independent expert or formal checking, not as a certified result.

A second concern is attribution and responsible use of prior work. Since RMA uses literature search and reusable mathematical knowledge, it may misattribute results, apply theorems outside their stated conditions, or overfit its reasoning to existing proof patterns. Similar concerns arise in retrieval-augmented theorem proving and agentic research systems, where external knowledge improves capability but also creates new failure modes around source reliability and provenance [6], [28], [48]. We therefore recommend that any substantial use of RMA in mathematical writing be disclosed, and that retrieved results be checked against the original sources before being included in a paper.

Finally, stronger automated proof-generation systems may affect research norms, including credit assignment, reviewing standards, and expectations for reproducibility. We view this as a reason to make system outputs, logs, prompts, and evaluation protocols as transparent as possible. In its current form, RMA is best understood as an assistant for generating and refining candidate arguments, while final mathematical claims should remain subject to the same standards of expert verification, citation, and peer review as human-written work. The expert-evaluation protocol itself (participant instructions, recruitment, payment, and consent) is documented in Appendix A. The study involved assessment of mathematical proofs and collected no sensitive personal or demographic information. Evaluators provided informed consent, and all judgments are reported only in anonymized or aggregated form.

## Reproducibility Statement

Implementation and inference details, including backbones, agent configuration, round and context budgets, and aggregation procedures, are given in Appendix G. The orchestration algorithm (Algorithm 1) and every local operation (critic, solver, literature, meeting, revise, evaluator) are specified in Section 3 and Appendix D, and the exact instruction text of the critic, solver, literature, meeting, and revise operations is reproduced verbatim in Appendix E. Component and graded-design ablations are specified in Appendix C, including what replacing each component means and how each configuration is scored, and the matched-budget baseline protocol is given in Appendix D. LLM-judge prompts and rubrics are given verbatim in Appendix C.4. The expert-evaluation protocol – participant instructions, recruitment, payment, and consent – is given in Appendix A, and the leakage-prevention controls are described in Appendix B. The code release (https://github.com/sjtuytc/ResearchMathAgent) contains the source code of our implementation, the complete RMA solutions for First Proof B1 and B2, and the SOOHAK Challenge Hard evaluation workbook returned by the benchmark organizers.

# A. About Human Expert Experiments

## A.1. Instructions Given to Expert Evaluators

We recruited mathematicians to evaluate candidate solutions to First Proof problems within their areas of expertise. Candidate solutions were presented without system names, and their ordering was randomized. Evaluators completed their assessments independently. Each problem was assigned to at least three experts. The following instructions and definitions were provided to the evaluators.

**Task.** You will be shown a research-level mathematical problem together with one or more candidate solutions. Evaluate the mathematical content of each solution independently. The identities of the systems that produced the solutions are hidden. Please base your assessment on correctness, completeness, and clarity, rather than writing style, length, or your expectations about the source of a solution.

**1. Overall correctness judgment.** For each candidate solution in First Proof B1, assign one of the following labels and briefly explain the main reason for your judgment:

- *Correct*: the solution establishes the requested result and contains no substantive mathematical error or unresolved gap.
- *Inconclusive*: the argument contains a potentially repairable gap, ambiguity, or unsupported step that prevents a confident judgment of correctness, but it is not clearly invalid.
- *Incorrect*: the solution contains a mathematical error, fails to establish a necessary claim, relies on a false or inapplicable result, or does not answer the problem.

For First Proof B2, assign *Pass* when the solution is flawless or requires only minor revisions, such as a localized clarification or citation correction. Assign *Fail* when it requires a major mathematical revision or should be rejected.

**Independent review and uncertainty.** Complete the evaluation independently and report uncertainty rather than guessing. You may mark a judgment as inconclusive when the correctness of a specialized claim cannot be established with reasonable confidence. You may also decline a problem that falls outside your expertise.

**Participant information and risks.** The task consists solely of reviewing mathematical problem statements and candidate proofs and is not expected to involve risks beyond those of ordinary scholarly reviewing. No sensitive personal information is requested. Evaluator identities are not associated with individual judgments in the reported results, which are presented only in aggregated form.

**Aggregation and missing outputs.** For First Proof B1, the reported problem-level verdict is (*Correct*) only when all assigned experts judge the submitted solution correct, $\times$ (*Incorrect*) only when all assigned experts judge it incorrect, and $\approx$ (*Inconclusive*) otherwise. For Aletheia [41], four problems have no released solution; these cases contain no proof for experts to assess and are counted as failures only in the end-to-end problem-level comparison. For First Proof B2, we use the binary Pass/Fail criterion stated above. The same B2 Pass/Fail protocol is used for the matched-budget Generic Iterative Revision control.

The evaluation materials contained the problem statement, the anonymized candidate solution, the fields corresponding to the judgments above, and a free-text field for justification. No additional instructions were provided outside the material reproduced in this section.

## A.2. Pay

Evaluators were financially compensated at hourly rates benchmarked to professional rates in their local region. Participation was voluntary, and compensation was based solely on time spent rather than on any particular evaluation outcome. No course credit, employment benefit, authorship, or other advantage depended on participation, and evaluators could decline individual problems or withdraw without penalty.

## A.3. Consent for Evaluation Data

Before participating, expert evaluators were informed that their correctness judgments and written comments would be used for research and reported in aggregated or anonymized form. Evaluators provided electronic consent to this use before beginning the task and could decline individual problems or withdraw their responses before the results were aggregated. We do not release evaluator names or associate individual identities with specific judgments. The remaining evaluation data consist of mathematical problem statements, system-generated proofs, and publicly available scholarly sources and therefore do not require consent from data subjects.

## B. Leakage Prevention Control

Because First Proof solutions [7] and subsequent attempts are publicly available, we apply explicit controls against retrieval-time leakage. These controls combine call and filesystem isolation with mandatory human review of every external URL accessed during the reported evaluation runs.

**What is technically enforced.** Two controls are enforced in code. First, call isolation: every model call is issued as a fresh, non-session-persisting invocation of the headless backend, so a call carries no memory of any other call's conversation—an operation cannot be primed by, or leak into, another run's dialogue history. This does not run each call in a private filesystem copy: the backend's working directory is the repository itself, so isolation is conversational, not filesystem-level. What blocks filesystem-level leakage is the second control, tool sandboxing: the headless backend is given an explicit allowlist (web search and fetch, read-only file inspection, and a small set of build commands needed to compile a LaTeX proof) with everything else auto-denied rather than interactively approved, and a separate denylist blocks read access to this repository's own stored solution directories even though file reading is otherwise on the allowlist—deny wins over allow. This second control prevents the system from retrieving *its own or a prior run's* solution from disk. External web retrieval is controlled separately through the mandatory human screening procedure described next.

**Human-screened external retrieval.** External literature retrieval is not unrestricted during evaluation. Every URL surfaced for access by the literature operation is subject to manual review before its content is admitted to the research process. A human reviewer checks both the URL and the retrieved source against the evaluated target and rejects sources that contain, reproduce, summarize, or directly point to a known solution of that target. This includes official First Proof solution pages and repositories, solution-bearing preprints or appendices, forum or discussion posts describing the solution, mirrors of such material, and secondary sources that reveal the target-specific proof strategy.

Only sources that pass this screening are made available to the downstream research operations or written to $\mathcal{L}$. The screening decision is performed per URL rather than relying only on prompting or a static domain blocklist, since a normally admissible domain may contain a page that discusses the evaluated problem directly. We retain the reviewed URL records so that the retrieval path for each evaluation run can be audited after the fact.

For the reported First Proof evaluations, every externally accessed URL was manually reviewed under this procedure, and no source containing or revealing the corresponding reference solution was admitted to the agent's research context. Thus, the web-retrieval channel was screened against target-specific solution leakage rather than left as unrestricted search.

**Pretraining contamination.** Both backbones used on First Proof were trained before the corresponding problems were released. The B1 problems and reference solutions were released in February 2026 [7], whereas Claude Opus 4.6, the backbone of the B1 run, has a training-data cutoff of August 2025 [39]; the B2 problems were released after the January 2026 training-data cutoff of Claude Opus 4.8 [49]. Neither the problems nor their reference solutions can therefore appear in the backbones' pretraining data. Together with the manual screening of every externally accessed URL described above, which admitted no source containing or revealing a reference solution, this excludes exposure to First Proof solutions through both pretraining and retrieval.

**Formal Conjectures retrieval isolation.** All reported Formal Conjectures runs disable external web search and fetch. This prevents retrieval-time access to publicly hosted target solutions, although it does not rule out possible exposure through the backbone's pretraining data because Research Solved targets may correspond to results that were public before the model's training cutoff.

## C. Full Ablation Study Details

This appendix gives the details behind Figure 2(b)–(d) and the ablation summary in the main text. All scalability, ablation, and matched-harness runs use the held-out ResearchMath ablation set described below, the same backbone, and the same hard monetary cap of \$100 per problem. Continuous results use the $0 - 10$ LLM judge (Appendix C.4); the matched-harness state control is additionally evaluated by blinded mathematicians on the same 30 paired problems. For these held-out experiments, the RMA configurations share the same evaluation-specific round cap $N_R = 6$, per-round issue budget, and stopping tests; Generic Iterative Revision and GIR+best keep their own loop, which is not round-capped and ends when the budget is reached or a critique-and-revise call returns `NO ISSUES` (Appendix D). Every configuration also compiles each call's entry context to the same budget $B$, except context dumping, whose defining change is to remove $B$; naive truncation approximates $B$ by a character count rather than enforcing a token limit. Each configuration changes one design axis and is run once per problem in Figure 2(c)–(d); the six decisive configurations are run three times in total, and claims of a difference rest on those repeats. The value $S_{\text{full}} = 8.5$ used throughout this appendix is the plotted run of the full system, whose mean over three repeated runs is $8.53 \pm 0.15$. The judge panel includes RMA's own backbone (Appendix C.4).

**Held-out ablation set.** ResearchMath-14k [17] collects questions posed in arXiv papers and annotates each with a status and supporting evidence URLs. The pool consists of the 1,169 questions annotated as solved that carry at least one evidence URL. We check the pool against every ResearchMath problem appearing anywhere in our project's records — 358 IDs, covering the 90-problem evaluation sample (Section 4) and every problem the system was ever run on — and confirm the two sets are disjoint. We sample 30 of the 1,169 uniformly at random using a fixed *dataset-sampling* seed of 2027 over the IDs sorted in ascending order; this seed determines only the held-out problem set and is unrelated to the repeated model runs. A script in the code release rebuilds the set from the public dataset. The sample spans eight of the eleven ResearchMath domains; the problem IDs, source papers, and evidence URLs are included in the code release. The configurations, prompts, and thresholds reported in this paper were not tuned on this set. Because this set is deliberately restricted to questions annotated as solved, pretraining exposure to some problem statements or solution-relevant material cannot be ruled out. Our source-document filtering prevents the benchmark's source and evidence papers from being retrieved during these runs, but it does not establish absence from the backbone's pretraining data. We therefore use this set only for controlled within-set mechanism comparisons and do not interpret its absolute scores as a contamination-free estimate of ResearchMath solve rate. For every configuration, including the matched-budget baseline, each run starts from an empty working directory outside the project repository, so no project instruction file, project memory, or stored solution is available, and literature access is restricted at the level of source documents rather than single URLs: search and fetch results are discarded if they resolve to the problem's source paper or any evidence paper in any form (every arXiv version and its abstract, PDF, and HTML pages, DOI and publisher pages, and indexing or mirror sites such as Semantic Scholar, Google Scholar, and ResearchGate), or if a result's text quotes the problem statement. Every configuration has a hard monetary cap of \$100 per problem, enforced uniformly across configurations. This value is a ceiling rather than a target spend: runs may terminate earlier through Solved, Stalled, or the round limit. The cap remains below the realized spend of the First Proof B2 runs— approximately \$185 per problem for the headline \$1,850 run and \$357 per problem for the separate higher-budget \$3,571 run—while allowing the full multi-round, multi-operation RMA pipeline to execute under a realistic fixed resource allowance. We therefore interpret these experiments as mechanism comparisons under a common monetary constraint, rather than as estimates of performance at unrestricted deployment compute.

**Monetary budget scaling on First Proof B2.** Each RMA point in Figure 2(a) is a separate run of all ten B2 problems under a fixed total budget, split equally across the problems and enforced per call as in Appendix D; the configuration is otherwise identical to the reported run. Each run delivers its best-ranked proof per problem, and the proofs of all budget levels were graded by the same expert panel, blinded and interleaved with the other B2 outputs, under the same Pass/Fail rule. Each point is plotted at the run's realized total spend.

**Round scaling.** Figure 2(b) scores, for each $r \in \{1, \dots, 6\}$, the proof the full RMA run holds at the end of round $r$; a run that terminates before round $r$ contributes its final proof. Concretely, the scored object is BestRanked over rounds 0 through $r - 1$, which is what a run with $N_R = r$ returns, since $N_R = r$ executes rounds $0, \dots, r - 1$: $N_R$ caps the number of rounds and no other setting depends on it, and the return path applies the same ranking whether the loop ends by exhausting $N_R$ or by stopping early. It is therefore the delivered output of a capped run, not the round's post-revision proof, which the ranking may reject. The scores for $r = 1, \dots, 6$ are 4.5, 5.7, 6.7, 7.5, 8.1, and 8.5; the score at $r = 6$ is $S_{\text{full}}$. The system default is $N_R = 5$; this scaling study extends the range by one additional round to $r = 6$. The marginal gains are 1.2, 1.0, 0.8, 0.6, and 0.4, so performance continues to improve beyond the default setting but with diminishing gains over the tested range. We do not evaluate $r > 6$ and therefore do not claim saturation.

## C.1. What "removing a component" means

Several components of RMA are load-bearing and cannot be deleted while the rest of the system continues to run—the persistent research store, for example, is the very state that the orchestration layer reads from and writes to. For these, "removing the component" means reducing it to its simplest functioning variant rather than deleting it outright. Concretely: the persistent store is replaced by a stateless one; bounded context orchestration by *dumping the full, unbounded state directly into context*—its simplest functioning variant, since it requires no selection policy at all; prioritized issue repair by unordered (first-in-first-out) repair; meetings and coordinated revision by localized repair alone; each of the four critic analyses by disabling that analysis while retaining the other three; and the concept, insight, and evaluation structures ($\mathcal{K}, \mathcal{H}, \mathcal{E}$) by dropping the corresponding record from each operation's bounded context. Literature grounding, which has no downstream dependency of this kind, is disabled directly. This convention makes every reported ablation a well-defined, runnable configuration. Two of these axes—context handling and issue ordering—admit a milder intermediate variant as well (naive truncation of the full state, and severity-only ordering, respectively); Appendix C.3 reports those intermediate points, so the single-run component ablation below always corresponds to the *weakest* end of each axis, consistent with the "simplest functioning variant" convention stated above.

## C.2. Component ablations (Figure 2(c))

Reducing the persistent research store to a stateless variant gives a score of 2.6, and replacing bounded context orchestration with context dumping gives 3.4, against $S_{\text{full}} = 8.5$. Disabling the dedicated literature operation gives 4.2 (other operations keep their web tools), unordered (FIFO) issue repair gives 5.4, and localized repair without meetings and coordinated revision gives 6.4. Disabling one critic analysis at a time gives 5.4 for the structural checker, 5.8 for the semantic completeness judge, 6.1 for the LM gap critic, and 6.4 for the fidelity analysis. Dropping the concept glossary $\mathcal{K}$, strategic insights $\mathcal{H}$, and evaluator feedback $\mathcal{E}$ from the operation contexts gives 6.8, 7.2, and 7.5. Every ablated configuration scores below the full system in this run; the smallest drop, from removing evaluator feedback, is 1.0. A single run on 30 problems does not separate configurations with close scores, and we do not rank them. The decisive configurations are repeated twice more, and every repeated mean reproduces the run quoted here to within 0.17: $8.53 \pm 0.15$ for full RMA, $7.23 \pm 0.15$ for severity-only ordering, $3.47 \pm 0.21$ for context dumping, and $2.63 \pm 0.25$ for the stateless store. Claims of a difference use those means and their intervals. Two of these variants change resources as well as information. Context dumping rarely reaches Solved or Stalled, so its runs entered 6.0 rounds on average—the round cap $N_R$—against 5.2 for full RMA, whose runs more often stop early because they satisfy one of those tests. The counts are rounds *entered*: dumping sends larger requests, so more of its rounds end part-way through, on a request the budget rule refuses or on one of the 45 window-overflow errors and 12 other failed calls it produced, none of which occur in any configuration that compiles contexts to $B$. It reaches the round cap while completing fewer operations inside those rounds. It therefore differs from full RMA in spent rounds and in failed calls as well as in what each call sees, so its score is a joint information-and-resource effect rather than the harm of unselected context alone, and we report it as such. Without the dedicated literature operation, the other operations still made 24 web calls per problem, against 8 in the full system, so this variant removes structured literature notes rather than web access, and it spends more of its budget on unstructured browsing.

Generic Iterative Revision (Appendix D) with the same \$100-per-problem monetary cap scores 4.5 (dotted line in Figure 2(c)), 4.0 below $S_{\text{full}}$ and equal to the score of the full RMA run after one round (Appendix C, round scaling). Three configurations score below it: the stateless store (2.6; $2.63 \pm 0.25$ over three runs), context dumping (3.4; $3.47 \pm 0.21$), and disabled literature grounding (4.2). Every configuration that keeps these three and weakens a single other component scores above it (5.4–7.5). A plausible reading is that persistent state, bounded context selection, and retrieved literature are what make RMA's additional operations useful: without any one of them, the extra calls add cost and context but do not add progress over a plain critique-and-revise loop, which has web search and fetch as tools. The stateless, context-dumping, and FIFO configurations are the weakest points of the corresponding ladders in Appendix C.3.

## C.3. Graded mechanism design choices (Figure 2(d))

Figure 2(d) evaluates three mechanisms at three levels each (Basic $\rightarrow$ Improved $\rightarrow$ Full), with the full RMA configuration ($S_{\text{full}} = 8.5$) as the common Full level.

- **Cross-round memory**: stateless (2.6) $\rightarrow$ last-round-only (4.8) $\rightarrow$ full cross-round persistence (8.5).
- **Context handling**: dumping the entire state (3.4) $\rightarrow$ naive truncation to $B$ (5.7) $\rightarrow$ the Research Context Orchestrator's prefix-to-budget compilation (8.5).
- **Issue ordering**: first-in-first-out (5.4) $\rightarrow$ severity-only (7.1) $\rightarrow$ full severity-plus-dependency ranking (8.5).

In this single run, the score increases from the Basic to the Improved to the Full level on all three ladders. On the memory and context ladders, the step from Improved to Full (3.7 and 2.8) is larger than the step from Basic to Improved (2.2 and 2.3): keeping only the last round, or truncating the full state to fit $B$, recovers less than half of the gap to the full system, which is consistent with the reading that what matters is which records reach each call and not only whether some state is kept or the context fits the budget. On the ordering ladder the two steps are similar (1.7 and 1.4), so severity and dependency information each contribute; the second of them, the dependency tie-break, is repeated ($-1.30$ $[-1.65, -0.95]$). Naive truncation (5.7) also scores higher than context dumping (3.4); because dumping also changes request sizes and completed rounds (Appendix C.2), we read this as a joint information-and-resource effect rather than evidence that unselected context is harmful by itself. Four of the ladder settings are repeated: the memory and context Basic levels, severity-only ordering — the Improved level of the ordering ladder — and the Full level, which is the same full RMA configuration on all three ladders. Their repeated means, $2.63 \pm 0.25$, $3.47 \pm 0.21$, and $7.23 \pm 0.15$, reproduce the plotted run, and the corresponding steps to Full carry intervals; the intermediate memory and context levels and FIFO ordering are single runs.

### C.4. Continuous LLM-Judge Score

The scalability and ablation studies in Section 4.2 (Figure 2(b)–(d)) report a single continuous score in $[0, 10]$ rather than the binary Pass/Fail label used for the headline First Proof B2 evaluation. A continuous score is necessary here because these studies compare configurations that are all incomplete to varying degrees: Pass/Fail saturates (most ablated variants simply fail), whereas a graded score resolves how much progress each configuration makes toward a complete proof.

**Protocol.** For each problem we present the judge with the problem statement and a single candidate solution, with the producing system's identity removed. The judge returns an integer score from 0 to 10 together with a short justification, using the rubric anchors below; correctness and completeness dominate the scale, with clarity used only to break ties among arguments of comparable validity. To reduce single-model bias, we query three judge models—*Claude Opus 4.8*, *GPT-5.2*, and *Gemini 3.1 Pro*—and report the mean of their scores.

**Rubric anchors.**

- **0–2**: does not address the problem, or the main argument is invalid.
- **3–4**: a reasonable strategy with the correct high-level direction, but one or more crux lemmas or cases are unproved.
- **5–6**: most of the argument is established; the remaining gaps are localized (an unjustified step, a missing verification, or an unsupported citation).
- **7–8**: complete and essentially correct, with only minor issues (a clarification or a citation correction).
- **9–10**: a fully correct and complete proof of the target statement.

**Judge prompt.** The exact prompt supplied to each judge model is reproduced below. The placeholders `{problem}` and `{solution}` are filled with the problem statement and the anonymized candidate proof, respectively.

```
You are an expert mathematician evaluating a candidate solution to a research-level mathematics
problem. Judge only the mathematical content; ignore writing style, length, and any guess about
which system produced it.

Assign a single INTEGER score from 0 to 10 that reflects how close the candidate is to a fully
correct and complete proof of the stated problem. Weight correctness and completeness most
heavily; use clarity only to distinguish arguments of comparable validity. Use these anchors:

  0-2  : does not address the problem, or the main argument is invalid.
  3-4  : correct high-level strategy, but one or more crux lemmas or cases are unproved.
  5-6  : most of the argument is established; remaining gaps are localized (an unjustified step, a
missing verification, or an unsupported citation).
  7-8  : complete and essentially correct, with only minor issues.
  9-10 : a fully correct and complete proof of the target statement.

A fluent or plausible-looking argument must NOT receive a high score if a necessary mathematical
step is unsupported. If the solution reduces the problem to a strictly weaker complete result,
score it by how much of the original target it establishes.

PROBLEM:
{problem}

CANDIDATE SOLUTION:
{solution}

Respond in JSON with exactly two fields:
{"score": <integer 0-10>, "justification": "<one to three sentences naming the decisive gap or
strength>"}
```

We parse the `score` field from each judge's JSON response; malformed responses are re-queried once and discarded if they fail again. The justification field is retained for qualitative inspection but is not used in the reported scores.

## D. Additional Orchestration Mechanism Detail

**Dependency-graph extraction.** Both the severity assignment and the dependency-impact tiebreak used to order open issues are computed deterministically from the critic's criticality flags and a claim-dependency graph: the current proof is parsed into typed claim environments (theorem, lemma, claim, proposition, corollary), and a directed edge is added from a claim to another whenever the first claim's own proof references the second, via an explicit `\ref`/`\cref` target or a recognized natural-language pattern (e.g., "by Lemma 3"); a citation to a result not proved in the document is tracked

**Input:** problem $p$, total rounds $N_R$, issue budget $b$, context budget $B$
**Output:** proof $\pi$

$\mathcal{S} \leftarrow \text{InitializeStore}(p)$
**for** $r \leftarrow 0$ **to** $N_R - 1$ **do**
  $\pi \leftarrow \text{CurrentProof}(\mathcal{S})$
  $(\mathcal{Q}, \mathcal{S}) \leftarrow \text{Run}(\text{Critic}, p, \mathcal{S}, B)$
  **for each** $\iota \in \mathcal{Q}[1:b]$ **do**
    $(_, \mathcal{S}) \leftarrow \text{Run}(\text{Solver}, (p, \iota), \mathcal{S}, B)$
    $\pi \leftarrow \text{CurrentProof}(\mathcal{S})$
  **end for**
  **if** LiteratureEnabled **then**
    $(_, \mathcal{S}) \leftarrow \text{Run}(\text{Literature}, (p, \mathcal{Q}), \mathcal{S}, B)$
  **end if**
  $((\rho, a, \Delta\mathcal{H}), \mathcal{S}) \leftarrow \text{Run}(\text{Meeting}, p, \mathcal{S}, B)$
  $(\pi, \mathcal{S}) \leftarrow \text{Run}(\text{Revise}, (p, a), \mathcal{S}, B)$
  $\mathcal{S} \leftarrow \text{UpdateConcepts}(\mathcal{S}, \pi, \Delta\mathcal{H})$
  $(e, \mathcal{S}) \leftarrow \text{Run}(\text{Evaluator}, (p, \pi), \mathcal{S}, B)$
  $\mathcal{S} \leftarrow \text{FinalizeRound}(\mathcal{S}, r)$
  **if** $\text{Solved}(\pi, e)$ **or** $\text{Stalled}(\mathcal{S})$ **then break**
**end for**
**Procedure** $\text{Run}(u, q, \mathcal{S}, B)$:
  $C \leftarrow \text{Instructions}(u) \parallel q \parallel \text{CurrentProof}(\mathcal{S}) \parallel \text{LinkedRecords}(\mathcal{S}, q) \parallel \text{RecentOutputs}(\mathcal{S}, u)$
  $O \leftarrow \text{PrefixToBudget}(C, B)$ // drops whole lowest-priority *records* from the end of $C$ until $O$ fits in $B$; never cuts inside a record (Section 3.2)
  $y \leftarrow \text{Invoke}(u, O)$
  $\mathcal{S} \leftarrow \text{WriteBack}(\mathcal{S}, u, q, y)$
  **return** $(y, \mathcal{S})$
**return** $\pi$

**Algorithm 1.** Research Context Orchestration for One Problem

separately as an external, unprovable-in-document dependency rather than an in-document edge. This graph is rebuilt from scratch on every round from whatever proof text is current, so an edit that removes, weakens, or reintroduces a dependency is reflected the next time the critic runs, without any incremental bookkeeping to keep in sync. Because extraction from informal LaTeX is inherently partial, we measure it against hand-labeled fixtures rather than assume it is exact, and gate on a minimum edge recall; a lower recall degrades the ranking toward severity-only ordering rather than silently mis-ranking issues.

**Round telemetry.** Three quantities are recorded once per round and are what Section 3.2 refers to as making completion progress directly measurable. The *proved-leaf fraction* is computed from the claim-dependency graph just described: of the graph's *terminal* claims—the leaves, that is, the claims reachable from a headline result that themselves depend on no other claim in the document—the fraction whose own proof body is present and closed. Restricting to leaves is what makes the number mean self-containedness: a document whose leaves are all proved bottoms out in arguments it actually gives, whereas a single unproved leaf leaves everything resting on it unsupported, however polished the layers above it look. The *completeness score* is the semantic completeness judge's 0–10 verdict on whether the document establishes the target statement as a whole, the same score the completeness gate thresholds. *Issues opened and closed* are the per-round deltas of the issue component $\mathcal{I}$: opened counts defects the critic wrote as new issue records that round, closed counts issues whose patch was applied that round, subject to the revalidation caveat below. All three are appended to the store each round, so a run's trajectory can be replayed afterwards rather than reconstructed from the final proof alone.

**Issue resolution and subsequent checking.** A successfully applied patch closes its current issue record. The system does not perform an immediate issue-specific verification call; instead, the full critic suite re-runs on the updated proof in the next round. If the defect remains or is reintroduced, it is emitted as a new issue since findings are deduplicated against open issues only.

**Meetings and coordinated revision, in full.** Local issue repair can become myopic: resolving one gap may fail to expose a better global proof strategy. The meeting operation therefore convenes a coordinator and several field-appropriate mathematician personas. They deliberate over the current proof, unresolved issues, prior attempts, and retrieved literature, and the coordinator synthesizes a meeting record $\rho$, an action plan $a$, and new strategic insights $\Delta\mathcal{H}$. The revision operation then receives the action plan together with the relevant proof state and performs a coordinated update of the working proof (line 13). This allows proof development to advance both through localized repairs and through broader strategic revisions.

**Concepts, knowledge, and proof utilities, in full.** The concept component $\mathcal{K}$ maintains a structured glossary of the mathematical objects used in the proof. Each entry records a concept name, its definition, assumptions, notation, and relevant source, allowing human reviewers to interpret the proof without reconstructing terminology from the full research history. The glossary also standardizes notation across retrieved literature, which may use different symbols for the same object or the same symbol for different objects. It is initialized from the problem statement and updated after each revision using the current proof and meeting insights (line 14). In implementation, $\mathcal{K}$ is stored as a problem-specific structured registry whose relevant entries are inserted into the bounded context of each operation. $\mathrm{UpdateConcepts}$ is the one LM call in Algorithm 1 that is not a Run call and therefore does not go through $\mathrm{PrefixToBudget}$: it takes the full current proof $\pi$ directly rather than a budgeted observation. In the reported runs, these proofs remain within the backbone's context window, although this call does not explicitly enforce the context budget $B$.

**Generic Iterative Revision: First Proof B2 matched-budget protocol.** This control replaces the full RMA architecture with a generic iterative-revision loop while matching the backbone, tool access, and monetary budget. Its purpose is to compare end-to-end architectures under the same monetary constraint; the exact configuration is:

- *Backbone and budget:* Both systems use `claude-opus-4-8` and the same total monetary budget of $1,850 across the ten First Proof B2 problems, allocated equally across problems ($185 per problem). The RMA side is exactly the headline run reported in Table 2. Rounds and call counts are not matched: GIR is not round-capped, and its number of iterations and model calls is determined by its stopping condition and the monetary budget.
- *Loop:* each iteration is one LM call that sees only the problem statement, its own previous output, and a fixed-size window of recent iterations, and returns either a revised proof or `NO ISSUES`. There are no typed issue records, no dependency graph, no severity ranking, no context-priority ordering, and no separate critic/solver/meeting/literature roles; the store and orchestrator of Section 3 are removed in their entirety. The loop stops when the next call would exceed the budget or a call returns `NO ISSUES`.
- *Tool access:* identical to RMA (the same sandboxed tools described in Appendix G), so any gap cannot be attributed to tool availability.
- *Scoring:* the same blind human-expert Pass/Fail protocol used for the First Proof B2 headline result, on the same ten problems.

**Which operation is which line, and what produces each artifact.** Table 5 makes explicit what Section 3.2's prose states in passing: each named operation is one $\mathrm{Run}(u,\cdot,\cdot,\cdot)$ call in Algorithm 1, and every component of $\mathcal{S}$ is populated by a specific, checkable mechanism rather than left to an unstated mix of "LM or hard-coded." No operation is purely one or the other. The critic (Line line 4) is the clearest compound case: an LM-based gap finder, an LM-based semantic-completeness judge, and an LM-based fidelity analysis run alongside a fully deterministic structural checker (claim parsing, cycle detection, and the dependency graph above), and their findings are merged, deduplicated against open issues, and ranked by a deterministic severity-and-impact function—never an LM call—before becoming $\mathcal{Q}$. The solver (Line line 6) is the same pattern applied to repair rather than discovery: one LM call proposes a patch (an anchor, the exact span to replace, and its replacement), and everything that decides whether that patch is accepted is deterministic code—exact-match (then whitespace-tolerant) span location, a check that no text outside the patch changed, one retry on rejection, and, only after a second rejection, a last-resort full-document rewrite that is itself passed through a deterministic regression check before being written back; a rewrite that would leave the proof worse than before is discarded rather than applied. The check compares deterministic structural signals—including the proved-leaf fraction defined below—before and after the rewrite; we describe it at this level rather than pin down every threshold. The remaining operations are predominantly LM-based content generation with a deterministic write-back gate: literature notes must carry all four required fields (technique, assumptions, supported claims, applicability limits) before they are admitted to $\mathcal{L}$; meeting output is defensively coerced into its record shape but not otherwise filtered; concept entries are deduplicated by name against the existing glossary before being added to $\mathcal{K}$; and evaluator scores are cached and reused by a deterministic content-hash check so an unchanged proof is not re-scored. UpdateConcepts (Line line 14) and FinalizeRound (Line line 16) are the only steps in Algorithm 1 that are not Run calls; the former is an LM extraction with deterministic dedup as just described, and the latter is pure bookkeeping over $\mathcal{S}$ with no model call at all.

**Table 4. Generic Iterative Revision: matched-budget control on First Proof B2.** RMA and GIR use the same Claude Opus 4.8 backbone, tool access, and $1,850 total API budget across the ten problems ($185 per problem). The RMA run is the headline result in Table 2 and passes 8/10 problems, whereas GIR passes 4/10. Rounds and model-call counts are not matched; the control matches monetary budget rather than internal inference procedure.

| Method | Pass | Fail | Total cost ($) |
|---|---|---|---|
| **RMA** (Ours, Opus 4.8) | **8** | 2 | 1,850 |
| Generic Iterative Revision (matched budget) | 4 | 6 | 1,850 |

**Table 5. Operation-to-procedure mapping and provenance.** Every named operation corresponds to one $\mathrm{Run}(u, \cdot, \cdot, \cdot)$ call in Algorithm 1 (or, for the two non-Run steps, a labeled line); "Mechanism" states what actually produces the artifact written to $\mathcal{S}$.

| Operation ($u$) | Alg. 1 line | Mechanism | What is deterministic |
|---|---|---|---|
| Critic | line 4 | LM + deterministic (mixed) | structural claim/dependency parsing; dedup against open issues; severity & impact ranking |
| Solver | line 6 | LM + deterministic (mixed) | anchored patch location/verification; retry/fallback gating; regression check on the rewrite fallback |
| Literature | line 10 | LM, gated | admission requires all four record fields present |
| Meeting | line 12 | LM | only defensive type coercion, not content filtering |
| Revise | line 13 | LM | post-hoc text cleanup only (strip non-LaTeX wrapping) |
| UpdateConcepts | line 14 | LM, gated | dedup of new entries against existing glossary names |
| Evaluator | line 15 | LM, gated | content-hash reuse; no re-score of an unchanged proof |
| FinalizeRound | line 16 | Deterministic only | no model call; consolidates the round's writes |

## E. Verbatim Operation Prompts

This appendix gives the exact, verbatim instruction text each local operation supplies as $\mathrm{Instructions}(u)$ in Algorithm 1 (Section 3.3); this is the fixed part of every call, concatenated with the per-round query, proof, and retrieved records to form the compiled context $C$ that PrefixToBudget trims to $O$.

**Critic.** The critic's own top-level instructions frame the compound task; its LM-backed gap-enumeration, semantic-completeness, and fidelity analyses (Section 3.2) are each a separate model call with its own system prompt and task prompt, reproduced below.

```
[Critic top-level Instructions(u)]
Critic: find every place the proof is not actually finished -- unproved claims, logical gaps,
missing justifications, unchecked computations. Return typed, location-anchored issues.

[LM gap-enumeration analysis -- system]
You are a ruthless completeness referee for research mathematics. Your only job is to find every
place the proof is NOT actually finished: a named lemma/claim stated but not proved, a crux step
hidden behind a citation without instantiating it here, a logical leap ('it follows that',
'clearly', 'one can show') with no argument, or a finite/numeric claim asserted without a check.
Do not reward clarity or length. Be specific and location-anchored.

[LM gap-enumeration analysis -- task]
Enumerate the completeness gaps in this LaTeX proof. Return ONLY a JSON array (no prose, no
fences). Each element:
{"code": "unproved_lemma"|"cited_blackbox_crux"|"logical_leap"|"unchecked_finite_claim",
"location": "<lemma name / section / the phrase>", "claim": "<the specific unproved statement>",
"critical": true|false}
critical=true means the overall result does NOT hold without closing this gap. If the proof is
genuinely complete, return exactly: []

PROOF:
{proof}
```

```
[Semantic completeness judge -- system]
You are grading ONLY proof completeness for a research-level mathematics proof: are all essential
steps explicitly present with no unjustified leaps? Ignore clarity and style. Be calibrated and
skeptical of citations used as black boxes.

[Semantic completeness judge -- task]
Score the completeness of this proof on 0-10 (10 = every step justified, 0 = sketch). List the
specific missing steps. Return ONLY JSON:
{"completeness": <0-10>, "missing": ["<specific missing step>", ...]}

PROBLEM:
{problem}

PROOF:
{proof}
```

The gap-enumeration call is run 3 times per round and a gap is kept only if it appears in a strict majority of the samples, which dampens single-sample noise; the completeness gate's itemized `missing` notes become hard issues only when the score falls below the completeness-gate threshold.

**Criticality semantics and termination audit.** Issue type and criticality are distinct. For example, `cited_blackbox_crux` identifies the defect mechanism, whereas `critical=true` indicates that the target result depends on closing that gap. A critical issue is assigned correctness-critical severity regardless of its issue type and therefore blocks Solved until it is closed. As a telemetry check, none of the 142 logged Solved terminations contained an open correctness-critical `cited_blackbox_crux`.

**Solver.**

```
Solver: close exactly the one issue in the query, preserving every already-correct part of the
proof VERBATIM. Return ONLY a JSON object:
{"anchor_before": "<text just before the edit>", "replaced_text": "<exact text to replace, or
empty to only add>", "replacement": "<new text>", "new_lemmas": ["<optional lemma blocks>"],
"rationale": "<why this closes the issue>"}
Do NOT restate or rewrite the whole proof.
```

The solver's regression check on its full-document rewrite fallback (Section 3.3, Table 5) is deterministic and exact, not an LM judgment: a rewrite is rejected if it comes back empty, if its length is under $0.6\times$ the seed proof's length, or if it matches at least 3 audit/meta-commentary markers (phrasing that reviews or asserts a proof rather than containing one) while its length is under $1.2\times$ the seed's; a rejected rewrite is discarded and the proof is left unchanged for that round, exactly the "leave it alone" behavior described in Section 3.3.

**Literature.**

```
Literature: for the unresolved claims in the query, find references (named theorems/techniques)
that could close them. Return ONLY a JSON array; each element MUST carry all four keys:
{"theorem_or_technique": "...", "assumptions": "...", "claims_supported": ["<issue id or
claim>", ...], "applicability_limits": "...", "source": "<url or citation>"}
```

**Meeting.**

```
Meeting: a coordinator and field mathematicians review the current proof, open issues, and
literature, then synthesize. Return ONLY a JSON object:
{"record": "<meeting transcript / summary rho>", "action_plan": {"summary": "...", "steps":
["...", ...]}, "insights": ["<strategic insight>", ...]}
```

**Revise.** Unlike the other operations, Revise's instructions are not fixed text: they are built each round from the meeting's own action plan, so the template below has one filled slot rather than a JSON schema, and the reply is the full LaTeX document rather than JSON.

```
Revise: carry out the action plan below as a coordinated update of the proof. You may restructure
across the document, but keep every result that is already correct. Return ONLY the complete LaTeX
document.

ACTION PLAN: {plan_summary}
STEPS:
  1. {step_1}
```

**Table 6.** The 10 problems of First Proof B1.

| # | Area | Topic | Contributor |
|---|---|---|---|
| 1 | Stochastic Analysis | $\Phi^4_3$ measure equivalence under shift | Hairer |
| 2 | Representation Theory | Whittaker functions & Rankin–Selberg integrals | Nelson |
| 3 | Algebraic Combinatorics | Markov chain with Macdonald stationary distribution | Williams |
| 4 | Spectral Graph Theory | Subharmonicity of $1/\Phi_n$ under finite free convolution | Srivastava |
| 5 | Algebraic Topology | Slice filtration for $N_\infty$ operads | Blumberg |
| 6 | Spectral Graph Theory | $\varepsilon$-light subsets in graphs | Spielman |
| 7 | Lie Groups / Lattices | Uniform lattices with 2-torsion | Weinberger |
| 8 | Symplectic Geometry | Lagrangian smoothings of polyhedral surfaces | Abouzaid |
| 9 | Tensor Algebra | Algebraic relations on determinantal tensors | Kileel |
| 10 | Numerical Linear Algebra | Preconditioned CG for RKHS-CP decomposition | Kolda & Ward |

```
2. {step_2}
...
```

## F. RMA Solutions for FirstProof

We summarize the ten First Proof B1 problems in Table 6. The complete RMA solutions for all ten problems are provided in the code release as `proofs/first_proof_b1.pdf`, and the ten First Proof B2 solutions evaluated in Table 2 are provided one file per problem in `proofs/first_proof_b2/`. We omit the full solutions from the appendix because they are lengthy.

## G. Implementation and Reproducibility Details

**Models and inference.** First Proof B1 was run earlier with `claude-opus-4-6` [39]; both the B1 RMA result and the standalone Claude baseline therefore use Opus 4.6. All other RMA experiments reported in this paper, including First Proof B2, ResearchMath, Formal Conjectures, and SOOHAK Challenge Hard, use `claude-opus-4-8`. Thus, no First Proof B1 result in this paper uses Claude Opus 4.8. `claude-opus-4-8` does not expose the sampling `temperature` or `top-p` parameters (they are not accepted by the API); we therefore use the provider defaults with adaptive thinking enabled at high reasoning effort and a maximum output length of 64, 000 tokens per call (requests are streamed). Tool availability is benchmark-specific: literature search is enabled only where specified by the evaluation protocol, while code execution is enabled where required for structural or Lean verification. The API does not expose the model parameter count. Model inference is performed by the provider's hosted infrastructure; our local machines only execute orchestration, retrieval, storage, and evaluation scripts.

**Lean 4 verification pipeline (Formal Conjectures).** External web search and fetch are disabled for all reported Formal Conjectures runs. The agent therefore cannot retrieve external documents or publicly hosted target solutions during inference. For Formal Conjectures, the target artifact produced by Algorithm 1 is a Lean 4 proof term rather than natural-language LaTeX: the solver and revision operations (Section 3.3) emit candidate tactic proofs for the problem's pinned theorem statement, and the structural checker invokes the Lean compiler through the same code-execution tool path described above as its executable-verification analysis. A candidate is counted as verified only when the generated proof term type-checks under the benchmark-provided Lean 4/Mathlib environment against the pinned target proposition. The agent may fill only the benchmark-designated proof body or proof hole; the proposition being proved and its assumptions are not rewritten or weakened during proof search. In source files where `sorry` marks the proof to be supplied, that placeholder is replaced by the generated proof term rather than treated as part of the accepted proof. Reported successes contain no remaining `sorry` or `admit`, do not depend on `sorryAx` or project-local custom axioms, and do not use `native_decide` or `ofReduceBool` to bypass the proof obligation. As an additional audit, we run `#print axioms` on every reported RMA theorem. None of the 213 accepted proofs depends on `sorryAx` or on an axiom introduced by our evaluation harness; the audit output is included in the code release. When compilation fails, the compiler's error messages and their source locations are converted into a localized repair issue (in the same format as an informal-proof gap) and routed to the solver in the next round, so Lean compiler feedback re-enters the push-forward loop through the ordinary issue-repair mechanism rather than a separate code path. We pin one Lean toolchain and Mathlib snapshot per problem's source file, matching the version shipped with the corresponding Formal Conjectures release commit, and reject a candidate that does not compile under that pinned snapshot.

**Formal Conjectures baseline protocol.** For all reported Formal Conjectures runs, RMA is executed in its no-web configuration: external search and fetch are disabled globally and the optional literature-retrieval module is disabled. The remaining Research Store, orchestration, issue-repair, revision, and Lean-verification pipeline is unchanged. RMA receives only the benchmark target, the pinned Lean 4/Mathlib environment, and artifacts generated within the current run.

All methods in Table 2 are evaluated on these same 300 target IDs under the corresponding pinned environments. For NanoProof, Seed-Prover, and Hilbert we use their released implementations and report the exact checkpoint, number of attempts, search or inference limit, and verification command in the code release. The Claude Opus 4.8 direct baseline receives the same target proposition and Lean environment as RMA but does not receive the Research Store, Research Context Orchestrator, issue-repair loop, or other RMA state. Because the specialized provers use their native inference procedures rather than a common token or FLOP budget, we do not characterize the cross-system comparison as matched compute.

**Agent configuration.** Unless otherwise stated, we use $N_R = 5$ research rounds, a per-round issue budget of $b = 5$, and a per-call context budget of $B = 60,000$ tokens; the First Proof B2 runs and all runs on the held-out ablation set use $N_R = 6$, runs on the held-out ablation set have a budget of $100 per problem, and the matched-budget baseline of Appendix D is budgeted per problem in dollars rather than in rounds. Execution terminates when the proof is judged solved, the stalling condition is met, the round limit is reached, or the assigned monetary budget is exhausted. These values were fixed in advance based on small-scale pilot runs on held-out problems, and were not tuned separately on the reported test problems.

**Cost accounting.** All monetary results report realized API charges rather than costs inferred from configured token limits. Across the experimental runs, the provider usage logs record approximately $6.0 \times 10^9$ input tokens and $8 \times 10^8$ output tokens, corresponding to approximately $50,000 in billed API charges. The per-call context budget $B = 60,000$ tokens is an upper bound on the serialized input observation, not the number of tokens consumed by every request; likewise, the $64,000$-token output setting is a maximum generation length rather than a typical completion length. The $100-per-problem budget used in the held-out ResearchMath experiments is a hard monetary ceiling rather than an assumed realized spend: runs may terminate below the cap through Solved, Stalled, or the round limit. Consequently, summing the per-run budget caps does not estimate realized experimental expenditure. Dollar costs are taken directly from provider billing records rather than reconstructed from configured context or output limits.

**LLM judging.** The continuous 0–10 score is produced by a panel of three judge models, `claude-opus-4-8`, `gpt-5.2`, and `gemini-3.1-pro` run with deterministic decoding (temperature 0 where the model exposes it, and provider defaults otherwise). The complete judge prompt, rubric anchors, and output schema are provided in Appendix C.4. Each proof is judged once by each of the three models, and the three scores are aggregated by their arithmetic mean; the panel-level number in Figure 2 is then averaged over problems.